\documentclass{article}
\usepackage{graphicx}

\usepackage[preprint,nonatbib]{neurips_2019}

\usepackage[utf8]{inputenc} 
\usepackage[T1]{fontenc}    
\usepackage{hyperref}       
\usepackage{url}            
\usepackage{booktabs}       
\usepackage{amsfonts}       
\usepackage{nicefrac}       
\usepackage{microtype}      

\usepackage[linesnumbered,ruled]{algorithm2e}

\title{Non-Determinism in Neural Networks for Adversarial Robustness}

\author{%
  Daanish Ali Khan \\
  Department of Computer Science\\
  Carnegie Mellon University\\
  Pittsburgh, PA 15213 \\
  \texttt{malikhan@andrew.cmu.edu} \\
  \And
    Linhong Li \\
  School of Computer Science\\
  Carnegie Mellon University\\
  Pittsburgh, PA 15213 \\
  \texttt{linhongl@andrew.cmu.edu} \\
  \And
    Ninghao Sha \\
  School of Computer Science\\
  Carnegie Mellon University\\
  Pittsburgh, PA 15213 \\
  \texttt{nsha@andrew.cmu.edu} \\
  \And
   Zhuoran (Oliver) Liu \\
  School of Computer Science\\
  Carnegie Mellon University\\
  Pittsburgh, PA 15213 \\
  \texttt{zhuoranl@andrew.cmu.edu} \\
  \And
    Abelino Jimenez \\
  Electrical and Computer Engineering\\
  Carnegie Mellon University\\
  Pittsburgh, PA 15213 \\
  \texttt{abjimenez@cmu.edu} \\
	\And
  Bhiksha Raj \\
  Language Technologies Institute\\
  Carnegie Mellon University\\
  Pittsburgh, PA 15213 \\
  \texttt{bhiksha@cs.cmu.edu} \\
  \And
   Rita Singh\\
  Language Technologies Institute\\
  Carnegie Mellon University\\
  Pittsburgh, PA 15213 \\
  \texttt{rsingh@cs.cmu.edu}
  }

\begin{document}

\maketitle

\begin{abstract}

Recent breakthroughs in the field of deep learning have led to advancements in a broad spectrum of tasks in computer vision, audio processing, natural language processing and other areas. In most instances where these tasks are deployed in real-world scenarios, the models used in them have been shown to be susceptible to adversarial attacks, making it imperative for us to address the challenge of their adversarial robustness.
Existing techniques for adversarial robustness fall into three broad categories: defensive distillation techniques, adversarial
training techniques, and randomized or non-deterministic model based techniques.  In this paper, we propose a novel neural network paradigm that falls under the category of randomized models for  adversarial robustness, but differs from all existing techniques under this category in that it models each parameter of the network as a statistical distribution with learnable parameters.
We show experimentally that this framework is highly robust to a variety of white-box and black-box adversarial attacks, while
preserving the task-specific performance of the traditional neural network model.

\end{abstract}

\section{Introduction}
\begin{figure}[h]
    \centering
    \includegraphics[width=0.5\columnwidth]{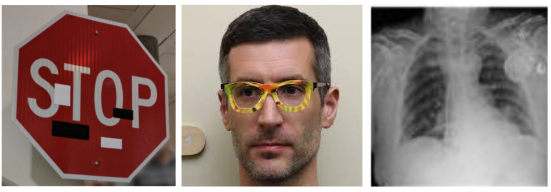}
    \caption{Examples of adversarial attaks on a variety of application domains. From left to right: Adversarial physical-world attack on stop signs \cite{adv-physical}, face-recognition systems \cite{adv-face}, and medical diagnosis systems  \cite{adv-medical}.  The adversarilly designed patches on the stop sign, spectacles on the face and noise in the x-ray all cause them to be misrecognized.}
    \label{fig:adv-examplesk}
\end{figure}

Neural systems are currently used in a broad spectrum of complex classification tasks, such as object recognition, speech processing, text generation etc. Many are deployed in large-scale tasks that are critical to human well-being and safety, such as biometric access points, medical assessments and self-driving cars. The systems themselves, however, are currently largely unprotected against malicious adversarial attacks  -- the presentation of inputs that have been purposely crafted to make the systems behave in incorrect ways \cite{Goodfellow2014} (Figure.\ref{fig:adv-examplesk}). 
Motivated largely by the desire to expose these vulnerabilities, a significant body of scientific literature has arisen in recent times on increasingly sophisticated techniques to generate adversarial instances -- inputs that may fool machine learning systems \cite{Goodfellow2014,DBLP:journals/corr/Moosavi-Dezfooli15,DBLP:journals/corr/NarodytskaK16}. 




With the increasing ubiquity of deep learning systems in the real world, the task of designing network architectures and learning paradigms that are robust to adversarial attacks is now recognized to be of paramount importance, and not surprisingly many solution approaches have been proposed in the literature \cite{defense-gan, towards-defense, defense-ensemble}. Adversarial training attempts to adjust classifier decision boundaries away from adversarial instances by including the latter in the training data \cite{DBLP:journals/corr/KurakinGB16a}. Distillation-based methods ``distil'' the trained networks into secondary networks to minimize their sensitivity to adversarial modifications of the input \cite{hinton2015distilling, DBLP:journals/corr/PapernotMWJS15}. Projection \cite{defense-project} and reconstruction methods \cite{defense-gan} attempt to project down (and possibly reconstruct) inputs prior to feeding them to the system, in order to eliminate adversarial modifications.  Randomization-based methods add noise or other random transformations \cite{defense-transform} to the input to mask out adversarial modifications \cite{Xie2017}. All of these methods assume the classification network itself to be deterministic. 

In this paper, we propose a novel and alternate route to adversarial robustness. In our approach the {\em parameters of the network are themselves stochastic}, having a statistical distribution with learnable parameters. Inference on the network too is stochastic. The premise behind our model is that the randomness in the model confounds the ability of the adversary to determine the minimal change of the input required to fool it. Randomness during inference also increases the probability of avoiding the increased-variance adversarial inputs that result.

We use a modified version of the Stochastic Delta Rule (SDR)\cite{SDR-original} to implement our stochastic models, employing a novel reparameterization trick to learn the distributions for the network parameters. We show experimentally that this framework does indeed provide protection against a variety of adversarial attacks in which other defences fail, while preserving the task-specific performance of the traditional neural network model. 

\section{Attacks: The Adversarial Threat Model}
We briefly discuss existing approaches to generate adversarial samples, to set the background for the discussion of current state-of-the-art defense techniques and our own proposal. 

The goal of the adversary is generally to generate examples that are perceptually indistinguishable from authentic (``clean'') inputs, but are incorrectly classified by the model. The adversary may either choose to generate  inputs that produce a specific  (bogus) output from the classifier ({\em targeted attack}), or modify a clean input such that it is classified as not belonging to its true class, without explicitly considering what it may be classified as instead ({\em untargeted attack}). We primarily consider the latter in this paper, although the proposed approach should generalize to targeted attacks as well.

Adversarial examples can be generated in either a white-box or a black-box setting. In the white-box scenario, the adversary has full access to the target model's architecture and gradients. Adversarial inputs are generally obtained by minimally perturbing clean inputs such that they now produce bogus outputs \cite{Goodfellow2014}. The perturbations are computed using variations of gradient descent. After a forward pass, gradients of an adversary-defined objective are back-propagated onto the clean input, revealing the adversarial perturbations that would confuse the model once added to it. Adversarial perturbations may be generated either by taking a single step along the gradient (\textit{one-step} methods) \cite{Goodfellow2014}  or taking steps iteratively until some stopping criterion is met (\textit{iterative} methods) \cite{DBLP:journals/corr/KurakinGB16a}. On the other hand, black-box adversarial attacks craft adversarial examples without any internal knowledge of the target network, which makes them much more applicable in the real-world setting. Here the general approach is to probe the classifier with inputs to obtain input-output pairs. These are used to learn how to generate adversarial samples \cite{Papernot2017}.

In our experiments, we choose to evaluate model robustness by classification accuracy under two white-box attacks, the Fast Gradient Sign Method (FGSM) and DeepFool Attack, and one black-box attack named LocalSearch, which we briefly review below. 

\subsection{White-box one-step attack example: Fast Gradient Sign Method}

Define the loss $L(X, y^{target})$ as a function of input $X$ and its target label $y^{target}$, which regular, non-adversarial training tries to minimize. To produce adversarial instances the adversary instead {\em increases} $L(X, y^{target})$, tweaking the input such that the model is less likely to classify it correctly. The \textbf{fast gradient sign method (FGSM)} proposed in \cite{Goodfellow2014} generates adversarial examples by taking a single step: 
\[X^{adv} = X + \epsilon * \texttt{sign}(\nabla_XL(X, y^{target}))\]
where the size of the perturbation $\epsilon$ is often subject to some restrictions \cite{DBLP:journals/corr/KurakinGB16a}. A common implementation of the FGSM attack is to gradually increase the magnitude of $\epsilon$ until the input is misclassified. Its iterative extension named \textbf{basic iterative method} has the following update rule \cite{KurakinGB16}:  
$$X^{adv}_0 = X, \qquad X^{adv}_{N+1} = Clip_{X, \epsilon}\left \{ X^{adv}_{N} + \alpha * sign(\nabla_XL(X^{adv}_{N}, y^{target})) \right \}$$ 
where $\alpha$ regulates the size of the update on each step and the total size of perturbation is capped at $\epsilon$ using $Clip_{X, \epsilon}$ \footnote{Here we borrow the notation from \cite{DBLP:journals/corr/KurakinGB16a}. $Clip_{X, \epsilon}(A)$ clips $A$ element-wise such that $A_{i,j} \in [X_{i,j} - \epsilon, X_{i,j} + \epsilon$].}. 

\subsection{White-box iterative attack example: DeepFool}

The \textbf{DeepFool} attack is an iterative attack similar to the basic iterative method but also takes into account the $\ell_2$-norm of the gradients when computing the update rule. The original paper suggests that the proposed method generates adversarial perturbations which are hardly perceptible, while the fast gradient sign method outputs a perturbation image with higher norm \cite{DBLP:journals/corr/Moosavi-Dezfooli15}. Such patterns are observed on both the MNIST and CIFAR-10 dataset using the state-of-the-art architectures. 

\begin{figure}[htb]
 \centering
\begin{minipage}{.6\linewidth}
\begin{algorithm}[H]
\SetAlgoLined
\caption{DeepFool Algorithm}
\KwResult{DeepFool Algorithm(binary case)}
{\bf Initialize:} $x_0\gets x,i\gets 0\;$

\While{$sign(f(x_i))=sign(f(x_0))$}{
  $r_i\gets \frac{f(x_i)}{||\nabla f(x_i)||_2^2}\nabla f(x_i)\;$
  
  $x_{i+1}\gets x_i + r_i\;$
  
  $i\gets i+1\;$ 
}
\end{algorithm}
\end{minipage}
\end{figure}



\subsection{Black-box iterative attack example: LocalSearch}


In the case of black-box attacks, the adversaries do not have access to model architecture and have no internal knowledge of the target network. These kinds of methods treat the network as an oracle and only assume that the output of the network can be observed on the probed inputs. The LocalSearch attack is accomplished by carefully constructing a small set of pixels to perturb by using the idea of greedy local search \cite{DBLP:journals/corr/NarodytskaK16}. This is an extension of a simple adversarial attack, which  randomly selects a single pixel and applies a strong perturbation to it in order to misclassify the input image. The LocalSearch attack is also an iterative procedure, where in each round a local neighborhood is used to refine the current image. This process minimizes the probability of assigning high confidence scores to the true class label, by the network. This approach identifies pixels with high saliency scores but without explicitly using any gradient information \cite{DBLP:journals/corr/NarodytskaK16}. 

\section{Defense: Methods Against Adversaries}
On the defenders' side, proposed measures against adversarial attacks include input validation and preprocessing, adversarial training, defensive distillation and architecture modifications. In the paragraphs below, we briefly review these methods and discuss how randomized training/models such as stochastic delta rule could increase model robustness against adversaries.

\subsection{Adversarial Training}
Adversarial training increases model robustness by providing adversarial examples to the model during training. The standard practice is to generate adversarial examples from a subset of the incoming batch of clean inputs dynamically. The model is then trained on the mixed batch of clean and adversarial inputs. However in order to do this, a specific method for generating adversarial examples must be assumed, preventing adversarial training from being adaptive to different attack methods. For example, \cite{DBLP:journals/corr/KurakinGB16a} showed that models adversarially trained using \textit{one-step} methods are fooled easily by adversarial examples generated using \textit{iterative} methods; models adversarially trained using a fixed $\epsilon$ could even fail to generalize to adversarial examples created using different $\epsilon$ values.

\subsection{Defensive Distillation}

Distillation was originally proposed in the context of model compression, aiming to transfer learned knowledge from larger, more complex models to more compact and computationally efficient models \cite{hinton2015distilling}. \textit{Defensive distillation} was first proposed by \cite{DBLP:journals/corr/PapernotMWJS15} as a training regime to increase model robustness against adversaries. 
The goal of \textit{defensive distillation} is not as much  transfer learning (for which distillation was originally proposed), but rather to train models to have smoother gradient surfaces with respect to the input -- such that small steps in the input space do not change the model's output significantly.

While smoothing out the gradients that adversaries usually use to create adversarial examples is effective in the setting described by the original paper, \cite{DBLP:journals/corr/CarliniW16} pointed out that the attack assumed by \cite{DBLP:journals/corr/PapernotMWJS15} (Papernot's attack) could be oblivious to  potentially stronger attacks. In addition, models trained with \textit{defensive distillation} possess no advantage against a modified version of Papernot's attack when compared to regularly trained models. Works such as \cite{Belagiannis2018} investigate the effect of network compression solely for the purpose of transferring model knowledge, but discovers the effect of robustness against adversaries as a side product.

\subsection{Randomized Methods \& Models}

Randomized training methods seek to improve the robustness of deep models by introducing randomness, irrespective of benign or adversarial samples, during the training process. For example, \cite{Xie2017} introduces a random resizing layer and/or zero-padding layer prior to the regular architectures of CNNs. Through experimental evaluation, the authors discovered that this method is particularly effective against iterative attacks, while other methods introduced above are better at handling single-step attacks. A combination of both methods, as the authors argue, achieve best performance against arbitrary adversaries. 


\section{Adversarial robustness through stochastic parameters}
Once trained, traditional neural networks have fixed parameters during inference.  This permits the adversary to obtain consistent responses from the system, as well as consistent gradient values required to compute perturbations.

In our approach the network parameters are themselves stochastic, drawn from a distribution. Each parameter $w_j$ in the network (which we assume without loss of generality to be a vector) has its own distribution $P(w_j; \theta_j)$ with parameter $\theta_j$.  When performing inference, the parameter value $w_j$ is drawn from its distribution, i.e. $w_j \sim P(w_j; \theta_j)$. 
The process of training the network comprises learning the parameters of the the distributions $\theta_j$, rather than the parameters $w_j$ themselves.

As a consequence of the stochasticity of the network, the gradients computed by an adversary (for the purpose of generating adversarial samples) will actually be stochastic, and may not generalize to other runs inference when the drawn parameter values are different. While it may seem that this effect should have little influence and average out in expectation, particularly for black-box attacks,  our experiments reveal that it is in fact sufficient to greatly decrease the efficacy of the adversary.

For our solution we use variants of the Stochastic Delta Rule \cite{SDR-original} to build our network. We describe these below. We provide the specifics of the original SDR training routine (\texttt{SDR-Decay}), along with our proposed fine-grained variant of SDR (\texttt{SDR-Learnable}). Along this trajectory, we will raise the issue of practical implementation concerns, the connection between SDR and regularization, and a qualitative explanation of feasibility of a SDR-augmented training routine in improving model generalizability and robustness against adversaries.

\subsection{The Stochastic Delta Rule}
First introduced in \cite{SDR-original}, SDR is revisited under the deep learning setting in \cite{Frazier-Logue2018}. During the forward pass of an SDR-equipped model,  parameters $w_j$ are not regarded as fixed values, but are rather random variables sampled from an arbitrary distribution $P(w_j; \theta_j)$ specified by parameters $\theta_j$. The choice of such distribution is arbitrary. For the purpose of our experiments, we assume that model parameters $w_j$ follow a normal distribution, {\em i.e.} $P(w_j; \theta_j) = N(\mu_j, \Sigma_j)$, and are independent of each other. The parameters $\mu_j$ and $\Sigma_j$ must be learned for each $w_j$.

In the discussions below on training these parameters, we will drop the subscript $j$ for brevity. Given a (minibatch of) training input(s) $(X,y)$ with features $X$ and labels $y$, at training iteration $t$, a model with the SDR training routine samples model parameters $w^{(t)} \sim N(\mu^{(t)}, \Sigma^{(t)})$, fits the current batch with respect to the sampled parameters, and performs the following updates:
\begin{eqnarray}\label{eq:sdr-decay}
    \mu^{(t+1)} &\gets& \mu^{(t)} - \alpha\nabla_{w} L(X, y, w^{(t)}) \nonumber \\
    \Sigma^{(t+1)} &\gets &  \Sigma^{(t)} + \beta|\nabla_{w} L(X, y, w^{(t)})|
\end{eqnarray}
where $\alpha, \beta$ are step sizes for the mean and variance respectively, and, as before, $L()$ is a loss function. 

On one hand, it can be readily observed that with $\beta = 0$ and zero-intialization of the covariance matrix, only the mean parameters are updated, and we recover the usual training routine without SDR. On the other hand, \cite{Frazier-Logue2018} states that if parameters are sampled from a binomial distribution with mean $Dp$ and variance $Dp(1-p)$, SDR could be regarded as a special case of dropout with probability $p$, only in this case the variance is not updated with respect to information gathered from the gradients. 

\subsection{SDR Update with Scheduled Variance Decay}

We first consider the SDR proposed by \cite{Frazier-Logue2018} which we formally present in algorithm \ref{alg:sdr-decay-alg0}. While updating model weights, the original SDR does not backpropagate gradients onto the parameter variances. Instead, the parameter variances are updated using the size of their associated means' gradients, $|\nabla_{w}\partial L(X, y, w)|$ (see Equation \ref{eq:sdr-decay}). As explained by \cite{Frazier-Logue2018} in the original paper, the intuition behind this update rule is that a larger gradient on the parameter mean would motivate expansion of the random node's distribution to explore potentially better weight values. Besides this update by gradient, the original SDR anneals the parameter variances by $\zeta$ to ensure asymptotically decaying variances (see Algorithm \ref{alg:sdr-decay-alg0} below). As training progresses, variances are shrunk so as to sample progressively concentrated parameters around the mean, for which reason we term this original SDR as \texttt{SDR-Decay}. In our implementation, we further introduce the decay schedule, $\tau$, to avoid over shrinkage and allow sufficient exploration of our model within the parameter space. At test time, under original the SDR, one would compute the forward pass directly with parameter means. However, in our case where SDR is applied to adversarial defense, doing so would defeat the purpose of generating stochastic gradients that mislead the adversary; thus we compute the forward pass of our model using weights sampled from the learned parameter distributions, as done during training.

\begin{algorithm}[H]\label{alg:sdr-decay-alg0}
 {\bf input} dataset $\{(X_i,y_i)\}_{i=1}^n$, decay schedule $\tau$, decay rate $\zeta \in (0,1]$, batch size $B$\\
 Initialize model parameters $\mu^{(0,0)}, \Sigma^{(0,0)}$ \\
 $\texttt{num-batches} \gets n // B$ \\
 \textbf{for} $e=1,2,\cdots$ until convergence: \\
 \qquad \textbf{for} $b=1,2,\cdots, \texttt{num-batches}$ \\
 \qquad \qquad Sample batch parameter weights $w \sim N(\mu, \Sigma)$ \\
 \qquad \qquad Compute forward pass of network with respect to $w$ \\
 \qquad \qquad Perform SDR parameter updates with respect to equations \ref{eq:sdr-decay} \\
 \qquad \qquad \textbf{If} $b \% \tau = 0$: $\Sigma \gets \zeta\Sigma$ \\
 \qquad \textbf{End for} \\
 \textbf{End for}
 \caption{\texttt{SDR-Decay}. The algorithm applies to every parameter in the network.}
\end{algorithm}

\subsection{SDR-Learnable}
Algorithm \ref{alg:sdr-decay-alg0} updates both parameter means and variances with $\nabla_{w}\partial L(X, y, w)$, requiring backward pass on only one set of parameters $w$ for each batch. We can, however, let both means and variances be fully learnable and have them updated with $\nabla_{w}\partial L(X, y, w)$ and $\nabla_{\Sigma}\partial L(X, y, w)$ respectively. The resulting algorithm, termed \texttt{SDR-Learnable}, produces a more realistic learning paradigm that approximates the behavior of variable parameters with higher fidelity. The drawback of such approach, of course, is doubling the computation required to compute the gradients. At test time, we sample parameters from learned means and variances, and perform inference on input data with sampled parameters. 

\begin{figure}[h]
    \centering
    \includegraphics[width=0.7\columnwidth]{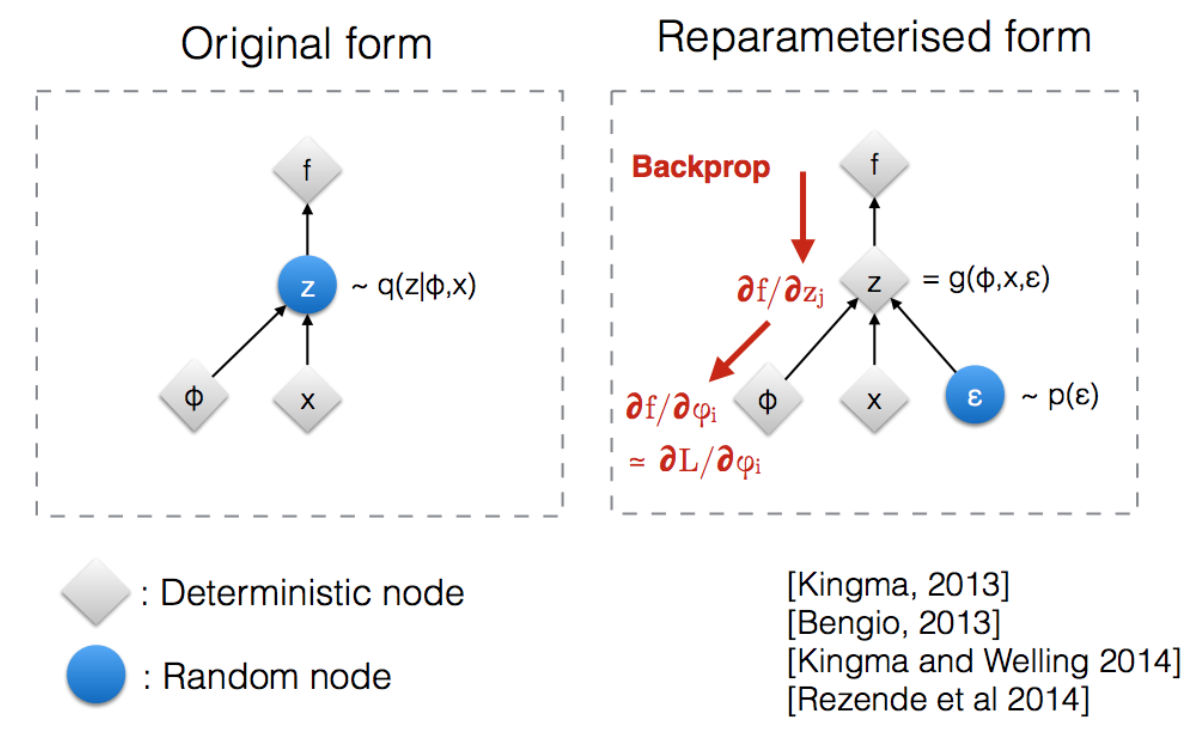}
    \caption{Reparameterization trick of \texttt{SDR-Learnable} update}
    \label{fig:reparam-trick}
\end{figure}

It is worth noting that with this formulation of the update rule, we require the loss function to be differentiable with respect to the parameters $\mu$ and $\Sigma$. However, the loss value is computed based on sampled realizations of many random variables along the forward pass. The sampling operation is not explicitly differentaible with respect to distribution parameters. In order to circumvent this issue, we use the reparametrization trick, as illustrated in figure \ref{fig:reparam-trick}, to make the network capable of backpropagating through random nodes. This technique essentially transfers the non-deterministic nature of the weight to another source of randomness, which then allows the randomly generated weights to be differentiable with respect to its parameters. 

In section \ref{var-shrink}, we show that the variance of the parameter distributions will shrink as the network is trained. Smaller variances result in less randomness in the forward pass, and as a result of this, the network does not exhibit consistent robustness to adversarial attacks. Furthermore, larger variances introduce a large amount of non-determinism in the forward pass. While this yields higher robustness, the overall task performance accuracy of the network diminishes.

In order to address this issue, we further modified the SDR architecture to use variance thresholds to ensure all parameter distributions' variances fall within a specified range during training. After performing each optimization step, we iterate through all parameters in the network and update them to fall within the desired range as shown in Algorithm \ref{alg:sdr-decay-alg} below.

\begin{algorithm}[H]\label{alg:sdr-decay-alg}
 \textbf{for} $\mu, \sigma$ in network parameters: \\
 \qquad \textbf{if} $\sigma >  \texttt{max var}$:\\
 \qquad \qquad $\sigma = \texttt{max var}$\\
 \qquad \textbf{else if} $\sigma <  \texttt{min var}$:\\
 \qquad \qquad $\sigma = \texttt{min var}$\\
 \textbf{End for}
 \caption{\texttt{Variance Threshold}}
\end{algorithm}

In order to find a trade-off between task-specific performance and adversarial robustness, we employed a training schedule to incrementally increase the variance of the parameter distributions. The minimum and maximum variances were initialized to $0.0$ and $1.0$ respectively. After every epoch of training, we evaluate the task-specific performance of the model on the test dataset. If the test performance begins to plateau, we increase both the minimum and maximum variances by $0.05$. This update occurs at most once every five epochs of training. As we increase the variance, the test accuracy decreases initially. After fixing the variance thresholds, the test accuracy improves again on the subsequent training iterations. Increasing the minimum variance allows us to improve the non-determinsm, which directly impacts the network's robustness to adversarial examples. By controlling the maximum variance, we ensure that the network is still capable of achieving high task-specific performance. Using this training schedule, we are able to train a network with high variance that performs well on the clean data, and is robust to adversarial attack.

\subsubsection{SDR-Learnable in Adversarial Learning}
The inference procedure of \texttt{SDR-Learnable} produces variable predictions with the same input, which motivates us to investigate its robustness against adversarial samples. Our qualitative motivation for this hypothesis is as follows: adversarial attacks are designed to lead models to misclassify, while inducing no human-recognizable changes to inputs. The adversarially modified and original clean inputs must be very similar. 
To account for such perturbations, the decision boundaries of a model must not only cater to the specific data points provided in the training data, but also to a {\em vicinity} of these points in the sample space. 

The variable treatment of parameters in \texttt{SDR-Learnable} is a step towards defensive strategies in two aspects: (1). the most effective gradient attack direction is computed with respect to one sampled parameter instance, and is less effective for another; (2) instead of fitting data at localized points, models with variable parameters create a decision region subject to the parameter distribution, hence allowing more robust prediction against adversarial samples. 

\subsubsection{SDR-Learnable in Generalization}
\label{var-shrink}
The description of decision regions, rather than spiky predictions, of \texttt{SDR-Learnable} naturally leads us to investigate its relationship to model generalization. In particular, we are interested in whether the inclusion of variable model parameters improves out-of-sample performance, and how do variances behave as training progresses. We present the following result as a first step towards the analysis.

Let $X \in \mathbb{R}^{N\times D}$ and $y \in \mathbb{R}^N$ be fixed, and $w \in \mathbb{R}^{D}$ be a random vector with $\mathbb{E}[w] = \mu$, $\mathrm{Cov}[w] = \Sigma$, then the risk of a linear regression model $\hat{y} = Xw$ takes the form
\begin{eqnarray}
\mathbb{E}_w\big[\| y-Xw\|^2\big] & = & \| y-X\mu\|^2 + \|X\Sigma^{1/2}\|^2
\end{eqnarray}

With simple algebra of expectation, we observe that
\begin{eqnarray}
    \mathbb{E}_w\big[\| y-Xw\|^2 \big] & =  &\vert\vert y\vert\vert^2 - 2y^T X\mathbb{E}_w[w] + \sum_{i=1}^N\mathbb{E}_w[(X_i^T w)^2] \\
    &= & \vert\vert y\vert\vert^2 - 2y^T X\mathbb{E}_w[w] + \sum_{i=1}^N \big(\mathrm{Var}[X_i^T w] + (X_i^T\mathbb{E}_w[w])^2\big) \\
    &= & \vert\vert y\vert\vert^2 - 2y^T X\mu + \sum_{i=1}^N (X_i^2\mu)^2 + \sum_{i=1}^N X_i^T\Sigma X_i \\
    &= &\vert\vert y-X\mu\vert\vert^2 + \vert\vert X\Sigma^{1/2}\vert\vert^2.
\end{eqnarray}

It can be readily observed that under the stylized linear regression model, \texttt{SDR-Learnable} is equivalent to regularizing parameter variances $\Sigma$ with penalty matrix $X$. In neural network training, we expect that the the Frobenius norm of $\Sigma$ to decay progressively, hence leading to more concentrated parameter samples. Notice that the decay step $\Sigma\gets \zeta\Sigma$ in \texttt{SDR-Decay} is a step towards artificial control of the magnitude of parameter variances, mimicking the behavior of \texttt{SDR-Learnable}.

It is worth noting that the parameter distribution does not necessarily lead to better generalization. Improved performance may be obtained by averaging the outcomes of multiple inferences, however this comes at the cost of adversarial robustness. Consequently we only perform a single pass of inference on any sample.

\section{Experiments \& Results}

We experimentally evaluate the efficacy of the proposed SDR-Learnable in model generalization and robustness against adversarial samples on the MNIST \cite{mnist} dataset, using the FoolBox toolkit \cite{foolbox} for adversarial samples. To compare performance, we evaluate three baseline models in addition to \texttt{SDR-Learnable}:  a standard 3-layer MLP, a 3-layer SDR MLP, and a 3-layer MLP with DropConnect $(p=0.2)$ \cite{pmlr-v28-wan13}. Dropconnect is a generalization of dropout \cite{dropout}, which introduces non-determinism in the network by randomly dropping network connections. This non-determinism results in an improvement in adversarial robustness.
The \texttt{SDR-Learnable} model used is a 3-layer MLP with learnable parameter distributions for weights and biases. All models had an input layer of size $784$, followed by three hidden layers of size $100$ and a final layer with $10$ output neurons, representing the class probabilities. The ReLU activation was used on all layers except the output layer which used the Softmax activation. To train all models, we used the Cross-Entropy Loss function.

Experimental evaluation was restricted to these simple models as gradient-based white-box attacks are significantly harder to defend against in this setting; larger and more complex models' gradients w.r.t the input are more dificult to estimate, adversely affecting the generated adversarial sample. Furthermore, black-box attacks will be able to estimate a simple models' decision boundaries to a higher degree of accuracy, resulting in more challenging adversarial inputs.



The models were trained using the standard train and test split on the MNIST dataset\cite{mnist}, the test-set results are reported in Table.\ref{tab:result}. The classification accuracy on the regular test-set was consistent across all models, with \texttt{SDR-Learnable} achieving an overall accuracy of $97.87\%$, outperformining the MLP and SDR baselines. It shows superior robustness against adversarial samples. Particularly in the case of the one-step FGSM attack, the \texttt{SDR-Learnable} model achieves a classification accuracy that is comparable to the classification accuracy  on the uncontaminated dataset.



\begin{table}[h]
    \small
    \centering
    \begin{tabular}{ |c|c|c|c|c| }
     \hline
      & Vanilla MLP & SDR-decay MLP &Learnable SDR MLP &  DropConnect  \\ 
     \hline
     Regular Samples & 97.59\% & 97.55\%& 97.87\% &  97.95\%\\ 
     \hline
     FGSM Attack & 0\% & 4.61\%& \textbf{94.86\%} &  45.9\% \\ 
     \hline
     DeepFool Attack & 0\% & 3.87\% & \textbf{78.42}\% &  45.48\%\\ 
     \hline
     LocalSearch Attack & 0\% &2.85\% & \textbf{88.89}\% &  26.53\%\\ 
     \hline
    \end{tabular}
    \caption{Model performance on regular test set, and under adversarial attack.}
    \label{tab:result}
    \vspace*{-1cm}
\end{table}

\section{Discussions and Conclusions}
From the results in Table.\ref{tab:result}, we note that \texttt{SDR-Learnable} is robust against one-step and iterative white-box attacks. The decrease in classification accuracy between the FGSM and DeepFool attacks is expected, as combating iterative attacks is a strictly harder task for adversarial defense\cite{obfuscated-gradients}. Without using the variance threshold schedule to train \texttt{SDR-Learnable}, the classification accuracy under the DeepFool attack was $40.15\%$. This indicates that the use of the variance threshold technique is crucial to defense against iterative white-box attacks. 
\texttt{SDR-Learnable} achieves an adversarial accuracy of $88.89\%$ against the iterative black-box attack LocalSearch. Without using the variance scheduler, the accuracy against LocalSearch is $44.05\%$. This indicates that there is significant benefit of using the variance scheduler to combat iterative black-box attacks. 

While \texttt{SDR-Learnable} has been shown to be robust against adversarial attacks, it was not trained using any adversarial examples. Existing work has shown that the use of adversarial examples during training results in an adversarially robust network; the same technique can be leveraged using \texttt{SDR-Learnable} networks to further improve their robustness.
Several existing techniques for adversarial defense can be incorporated into our network architecture, further improving model robustness.


In the reported experiments, \texttt{SDR-Learnable} was used only in an MLP, but the technique can be easily applied to almost any neural network architecture, resulting in at most twice the number of original parameters (each parameter is replaced with a distribution parameterized by a mean and a variance term). In our experiments, there was no significant difference in the time taken to train or evaluate the \texttt{SDR-Learnable} network. 

In conclusion, we have demonstrated that non-determinism in the model parameters improves robustness against white-box and black-box attacks.
The \texttt{SDR-Learnable} technique can be adapted using any network architecture, maintaining task-specific performance and providing defense against one-step and iterative whitebox and blackbox attacks, at no significant additional parameter cost. 
Furthermore, while iterative white-box attacks have been shown to compromise defense models based on stochastic gradients, we have shown that explicitly increasing the parameter variances while maintaining task-specific performance in \texttt{SDR-Learnable} significantly improves robustness.




\small
\bibliography{neurips}

\begin{thebibliography}{10}

\bibitem{Goodfellow2014}
Ian~J. {Goodfellow}, Jonathon {Shlens}, and Christian {Szegedy}.
\newblock {Explaining and Harnessing Adversarial Examples}.
\newblock {\em arXiv e-prints}, page arXiv:1412.6572, Dec 2014.

\bibitem{DBLP:journals/corr/Moosavi-Dezfooli15}
Seyed{-}Mohsen Moosavi{-}Dezfooli, Alhussein Fawzi, and Pascal Frossard.
\newblock Deepfool: a simple and accurate method to fool deep neural networks.
\newblock {\em CoRR}, abs/1511.04599, 2015.

\bibitem{DBLP:journals/corr/NarodytskaK16}
Nina Narodytska and Shiva~Prasad Kasiviswanathan.
\newblock Simple black-box adversarial perturbations for deep networks.
\newblock {\em CoRR}, abs/1612.06299, 2016.

\bibitem{defense-gan}
Pouya Samangouei, Maya Kabkab, and Rama Chellappa.
\newblock Defense-gan: Protecting classifiers against adversarial attacks using
  generative models.
\newblock {\em arXiv preprint arXiv:1805.06605}, 2018.

\bibitem{towards-defense}
Aleksander Madry, Aleksandar Makelov, Ludwig Schmidt, Dimitris Tsipras, and
  Adrian Vladu.
\newblock Towards deep learning models resistant to adversarial attacks.
\newblock {\em arXiv preprint arXiv:1706.06083}, 2017.

\bibitem{defense-ensemble}
Florian Tram{\`e}r, Alexey Kurakin, Nicolas Papernot, Ian Goodfellow, Dan
  Boneh, and Patrick McDaniel.
\newblock Ensemble adversarial training: Attacks and defenses.
\newblock {\em arXiv preprint arXiv:1705.07204}, 2017.

\bibitem{DBLP:journals/corr/KurakinGB16a}
Alexey Kurakin, Ian~J. Goodfellow, and Samy Bengio.
\newblock Adversarial machine learning at scale.
\newblock {\em CoRR}, abs/1611.01236, 2016.

\bibitem{hinton2015distilling}
Geoffrey Hinton, Oriol Vinyals, and Jeff Dean.
\newblock Distilling the knowledge in a neural network.
\newblock {\em arXiv preprint arXiv:1503.02531}, 2015.

\bibitem{DBLP:journals/corr/PapernotMWJS15}
Nicolas Papernot, Patrick~D. McDaniel, Xi~Wu, Somesh Jha, and Ananthram Swami.
\newblock Distillation as a defense to adversarial perturbations against deep
  neural networks.
\newblock {\em CoRR}, abs/1511.04508, 2015.

\bibitem{defense-project}
Yang Song, Taesup Kim, Sebastian Nowozin, Stefano Ermon, and Nate Kushman.
\newblock Pixeldefend: Leveraging generative models to understand and defend
  against adversarial examples.
\newblock In {\em International Conference on Learning Representations}, 2018.

\bibitem{defense-transform}
Chuan Guo, Mayank Rana, Moustapha Cisse, and Laurens van~der Maaten.
\newblock Countering adversarial images using input transformations.
\newblock {\em arXiv preprint arXiv:1711.00117}, 2017.

\bibitem{Xie2017}
et~al. Xie, Cihang.
\newblock Mitigating adversarial effects through randomization.
\newblock {\em \url{arXiv:1711.01991}}, 2017.

\bibitem{SDR-original}
Stephen~José Hanson.
\newblock "a stochastic version of the delta rule.
\newblock {\em Physica D: Nonlinear Phenomena 42.1-3 (1990): 265-272.}, 1990.

\bibitem{adv-physical}
Kevin Eykholt, Ivan Evtimov, Earlence Fernandes, Bo~Li, Amir Rahmati, Chaowei
  Xiao, Atul Prakash, Tadayoshi Kohno, and Dawn Song.
\newblock Robust physical-world attacks on deep learning models.
\newblock {\em arXiv preprint arXiv:1707.08945}, 2017.

\bibitem{adv-face}
Mahmood Sharif, Sruti Bhagavatula, Lujo Bauer, and Michael~K Reiter.
\newblock Accessorize to a crime: Real and stealthy attacks on state-of-the-art
  face recognition.
\newblock In {\em Proceedings of the 2016 ACM SIGSAC Conference on Computer and
  Communications Security}, pages 1528--1540. ACM, 2016.

\bibitem{adv-medical}
Samuel~G Finlayson, Hyung~Won Chung, Isaac~S Kohane, and Andrew~L Beam.
\newblock Adversarial attacks against medical deep learning systems.
\newblock {\em arXiv preprint arXiv:1804.05296}, 2018.

\bibitem{Papernot2017}
et~al. Papernot, Nicolas.
\newblock Practical black-box attacks against machine learning.
\newblock {\em Proceedings of the 2017 ACM on Asia Conference on Computer and
  Communications Security}, 2017.

\bibitem{KurakinGB16}
Alexey Kurakin, Ian~J. Goodfellow, and Samy Bengio.
\newblock Adversarial examples in the physical world.
\newblock {\em CoRR}, abs/1607.02533, 2016.

\bibitem{DBLP:journals/corr/CarliniW16}
Nicholas Carlini and David~A. Wagner.
\newblock Defensive distillation is not robust to adversarial examples.
\newblock {\em CoRR}, abs/1607.04311, 2016.

\bibitem{Belagiannis2018}
et~al. Belagiannis, Vasileios.
\newblock Adversarial network compression.
\newblock {\em \url{arXiv:1803.10750}}, 2018.

\bibitem{Frazier-Logue2018}
Noah Frazier-Logue and Stephen~José Hanson.
\newblock Dropout is a special case of the stochastic delta rule: faster and
  more accurate deep learning.
\newblock {\em \url{arXiv:1808.03578}}, 2018.

\bibitem{mnist}
Li~Deng.
\newblock The mnist database of handwritten digit images for machine learning
  research [best of the web].
\newblock {\em IEEE Signal Processing Magazine}, 29(6):141--142, 2012.

\bibitem{foolbox}
Jonas Rauber, Wieland Brendel, and Matthias Bethge.
\newblock Foolbox: A python toolbox to benchmark the robustness of machine
  learning models.
\newblock {\em arXiv preprint arXiv:1707.04131}, 2017.

\bibitem{pmlr-v28-wan13}
Li~Wan, Matthew Zeiler, Sixin Zhang, Yann~Le Cun, and Rob Fergus.
\newblock Regularization of neural networks using dropconnect.
\newblock In Sanjoy Dasgupta and David McAllester, editors, {\em Proceedings of
  the 30th International Conference on Machine Learning}, volume 28:3 of {\em
  Proceedings of Machine Learning Research}, pages 1058--1066, Atlanta,
  Georgia, USA, 17--19 Jun 2013. PMLR.

\bibitem{dropout}
Nitish Srivastava, Geoffrey Hinton, Alex Krizhevsky, Ilya Sutskever, and Ruslan
  Salakhutdinov.
\newblock Dropout: A simple way to prevent neural networks from overfitting.
\newblock {\em Journal of Machine Learning Research}, 15:1929--1958, 2014.

\bibitem{obfuscated-gradients}
Anish Athalye, Nicholas Carlini, and David Wagner.
\newblock Obfuscated gradients give a false sense of security: Circumventing
  defenses to adversarial examples.
\newblock In {\em Proceedings of the 35th International Conference on Machine
  Learning, {ICML} 2018}, July 2018.

\end{thebibliography}
\bibliographystyle{unsrt}




\end{document}